\documentclass[lettersize,journal]{IEEEtran}
\usepackage{amsmath,amsfonts}
\usepackage{array}
\usepackage[caption=false,font=normalsize,labelfont=sf,textfont=sf]{subfig}
\usepackage{textcomp}
\usepackage{stfloats}
\usepackage{url}
\usepackage{verbatim}
\usepackage{graphicx}
\usepackage{booktabs}
\usepackage{multirow}
\usepackage[numbers]{natbib}
\usepackage{pifont}
\usepackage{overpic}
\usepackage{footnote}

\usepackage[ruled,linesnumbered]{algorithm2e}
\usepackage{siunitx}
\usepackage{amssymb}
\usepackage[switch]{lineno}
\usepackage{hyperref}
\usepackage[capitalize]{cleveref}

\crefname{algorithm}{Algorithm}{Algorithms}
\crefname{figure}{Fig.}{Figs.}
\DeclareMathOperator*{\argmin}{arg\,min}

\begin{document}

\title{Cascaded Robust Rectification for Arbitrary Document Images}
\author{
    Chaoyun Wang,
    Quanxin Huang,
    I-Chao Shen,
    Takeo Igarashi,
    Nanning Zheng,~\IEEEmembership{Fellow,~IEEE,}
    Caigui Jiang
   \thanks{
    This work was supported by NSFC under Grant No. 62495092 and Natural Science Basic Research Plan in Shaanxi Province of China (No. 2025SYS-SYSZD-023), China Scholarship Council (No. 202406280284). \textit{(Corresponding author: Caigui Jiang.)}
    \par
    Chaoyun Wang, Quanxin Huang, Nanning Zheng, and Caigui Jiang are with the National Key Laboratory of Human-Machine Hybrid Augmented Intelligence, National Engineering Research Center for Visual Information and Applications, and Institute of Artificial Intelligence and Robotics, Xi’an Jiaotong University, Xi'an 710049, China. E-mail: \{chaoyunwang@stu.xjtu.edu.cn, qxhuang@stu.xjtu.edu.cn, nnzheng@mail.xjtu.edu.cn, cgjiang@xjtu.edu.cn\}.
    \par
    I-Chao Shen and Takeo Igarashi are with The University of Tokyo, Tokyo 113-8656, Japan. E-mail: \{jdilyshen@gmail.com, takeo@acm.org\}.
    \par
}
}

\maketitle

\begin{abstract}
Document rectification in real-world scenarios poses significant challenges due to extreme variations in camera perspectives and physical distortions. 
Driven by the insight that complex transformations can be decomposed and resolved progressively, we introduce a novel multi-stage framework that progressively reverses distinct distortion types in a coarse-to-fine manner.
Specifically, our framework first performs a global affine transformation to correct perspective distortions arising from the camera's viewpoint, then rectifies geometric deformations resulting from physical paper curling and folding, and finally employs a content-aware iterative process to eliminate fine-grained content distortions.
To address limitations in existing evaluation protocols, we also propose two enhanced metrics: layout-aligned OCR metrics (AED/ACER) for a stable assessment that decouples geometric rectification quality from the layout analysis errors of OCR engines, and masked AD/AAD (AD-M/AAD-M) tailored for accurately evaluating geometric distortions in documents with incomplete boundaries. Extensive experiments show that our method establishes new state-of-the-art performance on multiple challenging benchmarks, yielding a substantial reduction of 14.1\%--34.7\% in the AAD metric and demonstrating superior efficacy in real-world applications. The code will be publicly available at \url{https://github.com/chaoyunwang/ArbDR}.
\end{abstract}


\section{Introduction}
The ubiquity of smartphones has rendered the capture of physical documents for digital archiving~\cite{borges2008document}, recognition~\cite{li2022fast}, analysis~\cite{zhang2022multimodal,feng2025dolphin,feng2024docpedia}, and retrieval~\cite{duan2013towards} a routine task. However, this convenience is frequently compromised by geometric distortions inherent to the capture process. These distortions, stemming from physical paper deformations (e.g., curls and folds) coupled with unstable camera perspectives, introduce significant warping. Such artifacts pose a considerable challenge for downstream automated tasks by disrupting the document's intrinsic grid-like structure, thereby making document image rectification a critical and widely studied problem.

Early research in this domain primarily involved traditional methods that treated rectification as a 3D reconstruction problem. Initial attempts relied on specialized hardware, such as structured light systems~\cite{brown2001document,meng2014active} or stereo cameras~\cite{ulges2004document}, to model the 3D geometry for subsequent flattening. While accurate, the reliance on such equipment made these methods impractical for general use. Consequently, subsequent work shifted toward single-image rectification, leveraging strong geometric priors like document boundaries~\cite{brown2006geometric,koo2012segmentation} and text line curvature~\cite{kil2017robust,kim2015document} to infer the page's shape. Although this eliminated the need for special hardware, the dependence on specific assumptions limited their generalizability and robustness in diverse, real-world scenarios.

The advent of deep learning marked a paradigm shift, establishing data-driven methods as the mainstream approach due to their robustness and inference efficiency. These techniques train deep neural networks, such as Convolutional Neural Networks (CNNs)~\cite{li2021survey} or Transformers~\cite{vaswani2017attention}, on large-scale datasets to learn a direct mapping from a distorted image to its rectified version~\cite{ma2018docunet,verhoeven2023uvdoc,xue2022fourier,ma2022learning,das2019dewarpnet,li2023layout}. This geometric transformation is typically parameterized as a dense displacement field~\cite{das2019dewarpnet,li2023layout,feng2021doctr,feng2022geometric,feng2025docscanner,feng2023deep,ma2022learning,zhang2022marior,dai2023matadoc,li2023foreground,xie2020dewarping}, an intermediate 3D representation~\cite{ma2018docunet}, or a set of sparse control points~\cite{wang2025axis,verhoeven2023uvdoc,xie2021document,yu2024docreal}. Despite their success, most current research has focused on idealized scenarios: documents with complete boundaries and clean backgrounds. While recent methods like Marior~\cite{zhang2022marior}, DocTr++~\cite{feng2023deep}, and MataDoc~\cite{dai2023matadoc} have begun to address documents with incomplete boundaries, their performance often degrades in the presence of complex background interference. Consequently, achieving robust rectification of arbitrary in-the-wild document images remains an open challenge. To bridge this gap, we propose a unified framework designed to robustly address these complex rectification challenges.

\begin{figure}[!htbp]
    \centering
    \includegraphics[width=1.0\linewidth]{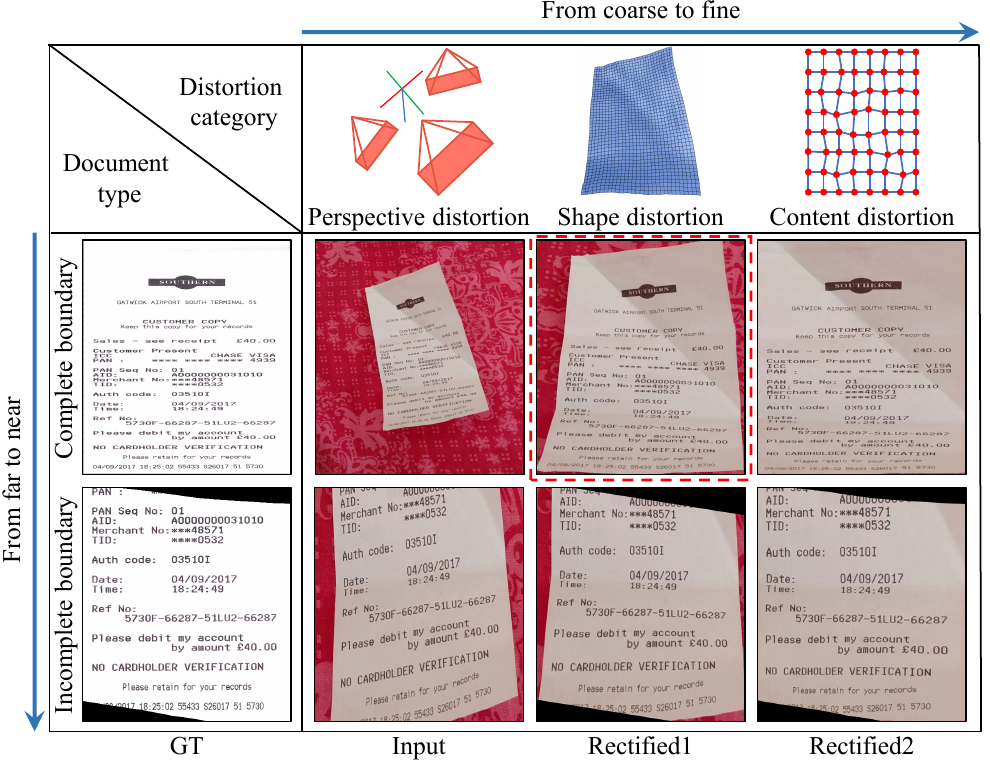}
    \caption{A systematic classification of arbitrarily distorted documents by type and distortion category. The highlighted red dotted box indicates the ideal cases targeted by current research.}
    \label{fig:introl}
\end{figure}

We systematically classify arbitrarily distorted document images along two primary axes: boundary completeness and distortion category, as illustrated in~\Cref{fig:introl}. Along the axis of document type, images are categorized based on the shooting perspective (from far to near) into complete-boundary documents and incomplete-boundary documents (which includes no-boundary cases). Along the axis of distortion category, we define a coarse-to-fine hierarchy of primary distortions based on their sources: 1) perspective distortion, caused by the camera's viewpoint and position; 2) geometric deformation, resulting from the physical bending or folding of the paper; and 3) content distortion, referring to fine-grained, local warps that become apparent in the final stage of rectification.

Existing methods primarily target the subset indicated by the dashed red box in~\Cref{fig:introl}---documents with complete boundaries and clean backgrounds. This focus limits their applicability in diverse, real-world scenarios. Our core proposal is to leverage this distortion classification to build a staged, coarse-to-fine framework. By sequentially addressing these distortion types in a coarse-to-fine process, we aim to create a unified solution for rectifying arbitrarily distorted document images.

Inspired by the classification of document distortions in~\Cref{fig:introl}, we propose an adaptive cascaded rectification framework to provide a unified solution for arbitrary document rectification. Our framework consists of three sequential stages: first, a localization network predicts control points that define a global affine transformation. This transformation removes background interference and corrects perspective distortion, normalizing the document into a canonical space. Next, a coarse rectification network corrects the document's overall geometric deformation. Finally, a fine-tuning network adaptively and iteratively eliminates local content distortions. 
While each stage operates independently, their transformations are cascaded to form a complete mapping from the distorted input to the rectified output, effectively implementing our coarse-to-fine strategy.

Experimental results demonstrate that our proposed method achieves state-of-the-art performance on several challenging public datasets. To the best of our knowledge, our method is the first to effectively rectify a wide range of document types and distortions, significantly outperforming previous leading methods and existing commercial applications. Furthermore, we introduce an enhanced evaluation metric for document rectification, which provides a more comprehensive and robust assessment of the final results.

Our main contributions are summarized as follows:
\begin{itemize}
    \item We propose a novel cascaded framework that progressively rectifies document images of arbitrary types, employing an adaptive iterative process to robustly handle a wide range of distortions from coarse to fine.
    \item We present two enhanced evaluation metrics to address limitations in existing protocols: the layout-aligned OCR metrics (AED/ACER) for robust evaluation, and masked AD/AAD (AD-M/AAD-M) for accurately evaluating documents with incomplete boundaries.
    \item Our method achieves state-of-the-art performance on multiple challenging benchmarks, outperforming prior work by a significant margin and demonstrating strong practical value.
\end{itemize}

\section{Related Work}
We survey traditional methods based on 3D reconstruction and 2D features, followed by deep learning paradigms such as dense field and sparse control point prediction. We conclude by analyzing the current trend of multi-stage, coarse-to-fine pipelines, highlighting the need for the systematic, integrated framework we propose.

\subsection{Traditional Rectification Methods}
Early approaches to document rectification were predominantly model-based and can be broadly classified into two categories: 3D surface reconstruction and 2D feature-based optimization.

\subsubsection{3D surface reconstruction}
These methods aimed to explicitly model the document's warped 3D surface. Some methods inferred this geometry from single-image cues like shading and texture~\cite{tan2005restoring,wada1997shape,zhang2009unified,liang2008geometric,he2013book}. Others constrained the problem by assuming the document was a specific parametric shape, such as a cylindrical~\cite{cao2003cylindrical} or developable surface~\cite{meng2014active,wang2024gso,WANG2026103970}, a concept later extended with isometric constraints~\cite{luo2022geometric,jiang2020quad}. However, these techniques were often slow and struggled to generalize, as their reliance on accurate feature detection made their optimization process susceptible to failure.

\subsubsection{2D feature-based optimization}
These methods bypassed 3D modeling, instead focusing on the geometric properties of 2D features like text lines and boundaries. These methods typically formulated rectification as an energy minimization problem, warping the image to satisfy geometric priors (e.g., forcing text lines to be straight and parallel)~\cite{ulges2005document,stamatopoulos2010goal,kim2015document,takezawa2017robust}. While later hybrid approaches used deep learning for more robust feature detection~\cite{jiang2022revisiting}, all these methods were fundamentally constrained by a common bottleneck: the reliable detection of low-level features is difficult in severely distorted images, thereby hindering their robustness and applicability.

\subsection{Deep Learning-Based Rectification}
The advent of deep learning has largely superseded traditional techniques by reframing rectification as an end-to-end image-to-image translation problem. These methods learn a geometric transformation from a distorted image to its planar counterpart and are typically classified into two paradigms: dense field prediction and sparse control point prediction.

\subsubsection{Dense Field Prediction}
The dominant approach predicts a dense, pixel-wise transformation field that can model complex, non-linear distortions with high fidelity. This field is typically represented in one of two ways:

\textbf{Forward Mapping.} The most intuitive approach predicts a forward map, defining where each pixel in the source image should move to in the rectified output. Pioneering this concept, DocUNet~\cite{ma2018docunet} used a U-Net to regress this map, with subsequent methods learning it at the patch~\cite{li2019document} or block level~\cite{li2023layout}. A similar technique predicts a displacement flow—a vector field of pixel offsets—as seen in works like Marior~\cite{zhang2022marior} and Xie et al.~\cite{xie2020dewarping}. However, forward mapping methods often require complex post-processing to resolve holes or overlaps in the output—a significant drawback.

\textbf{Backward Mapping.} To overcome these limitations, most modern methods predict a backward map. This approach inverts the mapping: for each pixel in the target grid, it predicts the corresponding coordinate to sample from in the source image. This formulation is directly compatible with differentiable grid sampling, making it highly efficient and the de facto standard. Popularized by DewarpNet~\cite{das2019dewarpnet}, this approach re-integrated 3D geometric reasoning, and has been adopted by numerous subsequent state-of-the-art methods~\cite{das2021end,feng2021doctr,feng2022geometric,feng2025docscanner,feng2023deep,dai2023matadoc,markovitz2020can}.

\subsubsection{Sparse Control Point Prediction}
As a lightweight alternative to dense field prediction, another line of research focuses on predicting a sparse grid of control points. This grid effectively serves as a compact representation of a backward map, offering greater computational efficiency and modeling flexibility~\cite{xie2021document,xue2022fourier,verhoeven2023uvdoc,yu2024docreal,wang2025axis}.

\subsection{Staged and Coarse-to-Fine Rectification}
Deep learning models excel on idealized inputs but often falter in in-the-wild scenarios characterized by background clutter, diverse perspectives, and incomplete objects. To address this, many researchers have adopted multi-stage, coarse-to-fine pipelines that decompose the complex rectification task into more manageable sub-problems~\cite{wu2022two,li2022accurate}.

The initial stage in these pipelines is a coarse correction to localize the document and remove the background. One popular strategy is to segment the foreground, eliminating background interference before the main rectification~\cite{dai2023matadoc,feng2021doctr,feng2022geometric,feng2025docscanner}. Other methods perform this correction using geometric transformations, such as those based on edge detection~\cite{zhang2022marior,yu2024docreal,ma2022learning} or Thin Plate Splines (TPS)~\cite{xue2022fourier}. A distinct recent approach involves an axis-alignment preprocessing step, which simplifies distortion without modifying the main network~\cite{wang2025axis}.

After the coarse correction, many methods employ an iterative refinement stage. For instance, DocScanner~\cite{feng2025docscanner} uses a recurrent architecture for progressive improvement, while Marior~\cite{zhang2022marior} predicts residual displacement fields to iteratively enhance the result. Other approaches leverage techniques like feature affine transformations for this fine-tuning step~\cite{liu2023rethinking}.

While these methods validate the coarse-to-fine paradigm, their stages often lack principled integration. In contrast, our work is founded on a principled taxonomy of document and distortion types (see~\Cref{fig:introl}). Building on this classification, we introduce a unified, cascaded framework that employs a staged and adaptive process for iterative content correction. This principled coarse-to-fine strategy provides a more systematic and robust solution, achieving superior performance and stability across diverse real-world documents.

\section{Approach}

Recognizing the inherent inefficiency of a single, monolithic model in handling the vast domain of document distortions, we employ a cascade of specialized models, each engineered to reverse a specific class of distortion in a progressive, coarse-to-fine manner. In this section, we first provide an overview of our framework, then detail the adaptive iterative condition and the loss functions used for training.

\subsection{Overall Pipeline}
\begin{figure*}[!htbp]
    \centering
    \includegraphics[width=1.0\linewidth]{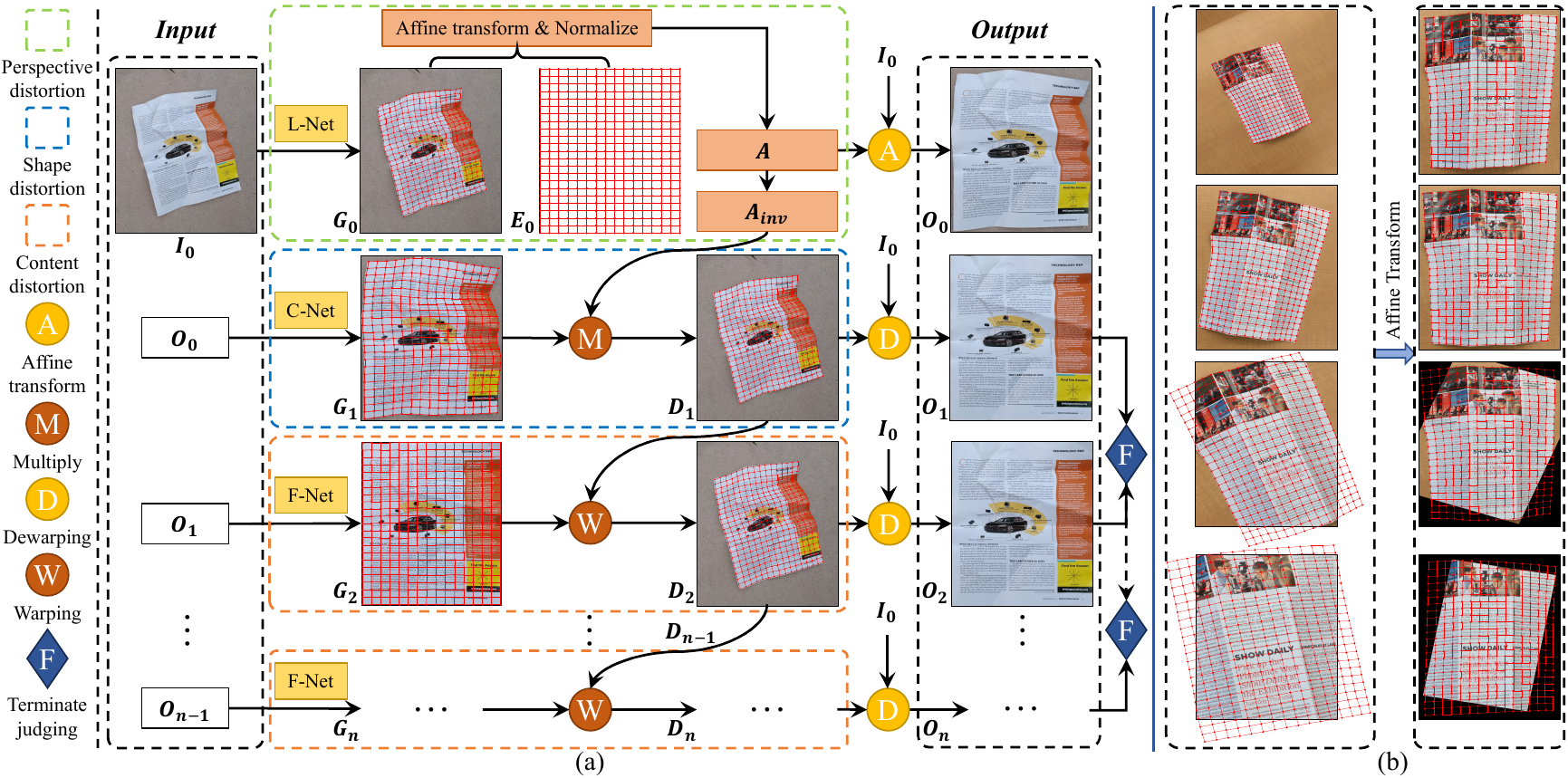}
    \caption{Architecture of the adaptive cascaded rectification framework. (a) Our pipeline reverses document distortions in a coarse-to-fine sequence: the L-Net corrects global perspective distortion, the C-Net rectifies coarse shape distortions, and the F-Net performs adaptive iterative refinement of content-level distortions. During inference, these specialist transformations are composed into a single backward mapping from the rectified output to the original input, ensuring high fidelity by minimizing resampling.
    (b) The effect of canonical view normalization. Our initial affine transformation converts diverse, challenging inputs (left) into a standardized domain (right), simplifying the task for subsequent networks.}
    \label{fig:pipeline}
\end{figure*}

Guided by our distortion taxonomy (\Cref{fig:introl}), our framework operationalizes this analysis as a three-stage cascade (\Cref{fig:pipeline}(a)). The architecture employs three expert networks: a Localization Network (L-Net) to reverse global perspective distortion, a Coarse Rectification Network (C-Net) to correct geometric deformations, and a Fine-tuning Network (F-Net) to resolve residual content warps.
The L-Net, C-Net, and F-Net share a unified architecture based on the backbone employed by Wang et al.~\cite{wang2025axis} and UVDoc~\cite{verhoeven2023uvdoc}. Specifically, we discard the auxiliary 3D coordinate branch employed in prior works to tailor the backbone for our cascaded pipeline.

\subsubsection{Perspective Distortion Rectification Stage}\label{subsec:perspective}
The first stage of our pipeline addresses perspective distortion caused by the camera's viewing angle. Given a distorted input image $I_0 \in \mathbb{R}^{H \times W \times 3}$, a localization network (L-Net) predicts a sparse control grid $G_0 \in \mathbb{R}^{h \times w \times 2}$ that captures the document's overall shape, as shown in \Cref{fig:pipeline}(a).

The final affine transformation $A$ is formulated as the composition of two matrices: an initial transform $A_{init}$ to correct the primary skew and a normalization transform $S_{norm}$ to center and scale the result. First, $A_{init}$ is estimated by solving a least-squares problem to map the predicted grid $G_0$ to a canonical, uniformly spaced grid $E_0$:
\begin{equation}
    A_{init} = \argmin_{A} \sum_{i} \left\| A(G_{0, i}) - E_{0, i} \right\|_2^2.
    \label{eq:affine_init}
\end{equation}
Subsequently, to standardize the output domain, the normalization transform $S_{norm}$ is computed to map the bounding box of the initially rectified grid, $A_{init}(G_0)$, into the normalized coordinate space $[-1, 1]^2$, including a safety margin $m$ to prevent over-cropping:

\begin{equation}
    S_{norm} = \mathcal{F}(A_{init}(G_0), m).
    \label{eq:snorm}
\end{equation}
The final affine transformation is the composition of these two operations, $A = S_{norm} \cdot A_{init}$. We then apply this matrix to the original image $I_0$ using the affine warping operator $\mathcal{W}_{A}$ to obtain the perspective-rectified output $O_0$:
\begin{equation}
    O_0 = \mathcal{W}_{A}(I_0, A).
    \label{eq:warp_affine}
\end{equation}
The primary benefit of this initial step is canonical view normalization, which simplifies the task for subsequent networks by substantially reducing the spatial variance of the distortions they must handle (\Cref{fig:pipeline}(b)).

\subsubsection{Shape Distortion Rectification Stage}
The second stage addresses non-linear geometric distortions, which result from physical deformation of the paper such as curls and folds. This is accomplished by a coarse rectification network (C-Net). Taking the normalized output from the first stage as input, the C-Net predicts a deformation field $G_1 \in \mathbb{R}^{h \times w \times 2}$. 
To obtain the final backward mapping grid $D_1$, we compose the predicted field $G_1$ with the inverse affine transformation $A_{inv}$ from the first stage. This composite grid is then used to warp the original image $I_0$ yielding the rectified output $O_1$:
\begin{equation}
	D_{1} = G_{1} \cdot A_{inv}, \qquad O_{1} = \mathcal{W}_{G}(I_0, D_{1}).
	\label{eq:warp_grid_cnet}
\end{equation}
This composite grid $D_1$ defines a backward mapping from the rectified output coordinates back to the original image $I_0$. The warping operator $\mathcal{W}_{G}$ then executes this mapping, using bilinear interpolation to sample the corresponding pixels.

\subsubsection{Content Distortion Rectification Stage}
The final stage of our pipeline performs fine-grained rectification to address subtle, content-level distortions, such as warped text lines, that may remain after the coarse correction. A Fine-tuning Network (F-Net) addresses this task by iteratively predicting a local deformation grid $G_{m+1} \in \mathbb{R}^{h \times w \times 2}$ at iteration $m$.

Unlike the initial composition with the global affine transform, the iterative refinement in this stage combines successive deformation grids using a resampling-based approach. The cumulative mapping grid from the previous step, $D_{m}$, is warped by the newly predicted grid, $G_{m+1}$. We denote this specialized composition, implemented via a \texttt{grid\_sample} operation, with the symbol $\triangleleft$. For each iteration $m \ge 1$, we compute the updated map and output as:
\begin{equation}
	D_{m+1} = D_{m} \triangleleft G_{m+1}, \qquad O_{m+1} = \mathcal{W}_{G}(I_0, D_{m+1}).
	\label{eq:warp_iterative_final}
\end{equation}

The operation $D_{m} \triangleleft G_{m+1}$ uses the local grid $G_{m+1}$ to look up and resample coordinates from the previous cumulative grid $D_{m}$, effectively chaining the non-linear backward mappings. The process is guided by an adaptive iterative strategy (detailed in \Cref{sec:iteractive}) that terminates the refinement when improvements plateau, optimizing both performance and efficiency.

\subsection{Iterative Stopping Condition}
\label{sec:iteractive}
\begin{figure}[!htbp]
    \centering
    \includegraphics[width=1.0\linewidth]{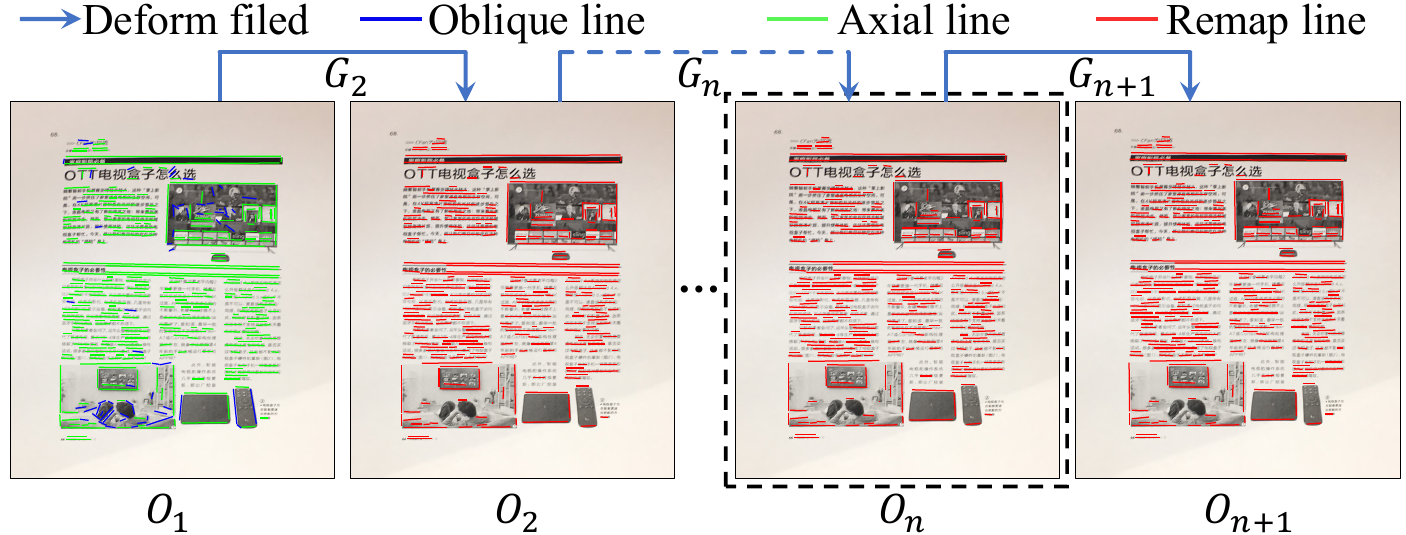}
    \caption{The geometric principle for our stopping condition. First, axis-aligned reference lines (green) are filtered from all detected lines (blue and green). These reference lines are then transformed by the iterative deformation field, and their final alignment is quantified as the line entropy score (Eq. \ref{eq:align_entropy}).}
    \label{fig:image_iter}
\end{figure}

Our adaptive stopping condition is guided by a core geometric principle from~\cite{wang2025axis}: leveraging the geometric prior that a rectified document typically exhibits textual and structural lines aligned with the canonical coordinate axes. To operationalize this principle in an unsupervised manner, we propose the Axis-Aligned Line Entropy ($H_{align}$), a metric where a lower value indicates better rectification. It is defined as the length-weighted average of the angular deviations of reference lines from the cardinal axes:
\begin{equation} \label{eq:align_entropy}
    H_{align} = \frac{\sum_{i=1}^{N} L_i \cdot \delta_i}{\sum_{i=1}^{N} L_i}.
\end{equation}
where $L_i$ is the length of the $i$-th line segment and $\delta_i$ is its acute angular deviation from the nearest axis. To establish a reliable set of reference lines for this metric, we apply the LSD algorithm~\cite{von2012lsd} to the coarse-rectified image $O_1$ (where primary geometric structures are restored) and then filter for nearly axis-aligned lines (e.g., within a $\theta_{thresh}$ of $5^{\circ}$), as illustrated in~\Cref{fig:image_iter}.

\begin{algorithm}[!htbp]
    \caption{Entropy-based Iterative Stopping}
    \label{alg:high_level_stop}
    \KwIn{Base image $O_1$,
    Max iteration count $M$,
    Threshold $\theta_{thresh}$,
    Improvement factor $\tau$.}
    \KwOut{Optimal iteration count $n_{opt}$}

    $L_{ref} \leftarrow \text{FilterAlignedLines}(\text{DetectLines}(O_1), \theta_{thresh})$\;
    $S_{best} \leftarrow \text{CalculateLineEntropy}(L_{ref})$\;
    $n_{opt} \leftarrow 0$; \quad $L_{current} \leftarrow L_{ref}$\;

    \For{$n \leftarrow 1$ \KwTo $M$}{
        $L_{warped} \leftarrow \text{WarpLines}(L_{current}, G_{n+1})$\;
        $S_{n} \leftarrow \text{CalculateLineEntropy}(L_{warped})$\;
        
        \uIf{$S_{n} < \tau \cdot S_{best}$}{$(S_{best}, n_{opt}, L_{current}) \leftarrow (S_{n}, n, L_{warped})$\;}
        
        \uElse{
            \KwSty{break}\;
        }
    }
    \KwResult{$n_{opt}$}
\end{algorithm}

The overall iterative process, detailed in Alg. \ref{alg:high_level_stop}, begins by calculating an initial entropy score from the reference lines in $O_1$. For each subsequent F-Net iteration, it transforms the reference lines and computes a new score. The process terminates if the alignment energy ($S_{n}$) fails to decrease significantly, indicating that the alignment quality has converged or degraded.

The \texttt{WarpLines} function updates the lines from the previous step by applying a transformation derived from Radial Basis Function (RBF) interpolation. This transformation is defined by the mapping from a canonical grid $U$ to the current deformation field $G_{n+1}$, and is applied to the input lines $L_{current}$ to produce the newly warped set $L_{warped}$:
\begin{equation}
    L_{warped} = \text{RBF}(U, G_{n+1}; L_{current}).
\end{equation}

\subsection{Loss Functions}
The network is trained with a composite loss function, formulated as follows:

\subsubsection{Component Losses}
To supervise the global alignment performed by the L-Net and C-Net, we introduce the affine transformation loss ($\mathcal{L}_{Affine}$). As detailed in~\Cref{subsec:perspective}, we compute the predicted affine matrix $A_{pred}$ from the predicted grid $G_{pred}$ and the ground truth matrix $A_{gt}$ from $G_{gt}$. The loss is the L1 distance between these matrices:
\begin{equation} \label{eq:affine_loss}
    \mathcal{L}_{Affine} = \| A_{pred} - A_{gt} \|_1.
\end{equation}
By directly supervising the six parameters of the transformation matrix, this loss provides a more stable and robust signal for learning global perspective correction than relying solely on grid point supervision.

Additionally, we incorporate three established loss terms inspired by Wang et al.~\cite{wang2025axis}: a 2D grid loss ($\mathcal{L}_{2D}$) on the L1 distance between control grids to ensure geometric accuracy, the SSIM loss ($\mathcal{L}_{SSIM}$) to maintain perceptual image quality, and an axis-alignment loss ($\mathcal{L}_{AL}$) as a regularizer to enforce the straightness of feature lines.

\subsubsection{Total Training Objectives}
These components are weighted and combined to form the final objectives for each network stage. Both the L-Net and C-Net are supervised by a comprehensive loss function that addresses global and coarse-level distortions:
\begin{equation}
    \mathcal{L}_{L/C-Net} = \alpha\mathcal{L}_{2D} + \beta\mathcal{L}_{SSIM} + \gamma\mathcal{L}_{AL} + \lambda\mathcal{L}_{Affine}.
    \label{eq:loss_lcnet}
\end{equation}
For the F-Net, which is dedicated to fine-grained refinement, the objective is modified. The affine loss ($\mathcal{L}_{Affine}$) is intentionally omitted, allowing the network to focus solely on correcting local, non-linear distortions without the global constraint already addressed by the preceding stages.
\begin{equation}
    \mathcal{L}_{F-Net} = \alpha\mathcal{L}_{2D} + \beta\mathcal{L}_{SSIM} + \gamma\mathcal{L}_{AL}.
    \label{eq:loss_fnet}
\end{equation}
Across all experiments, the weights are set as $\alpha=1, \beta=0.05, \gamma=0.2,$ and $\lambda=5$.

\section{DATASET AND METRICS}
Prior research has largely been confined to idealized document images with well-defined boundaries. In contrast, our work tackles the more realistic challenge of documents with arbitrary boundaries and complex backgrounds. This section therefore reviews the benchmarks for evaluation (\Cref{sec:benchmarks}) and introduces necessary enhancements to the evaluation protocols (\Cref{sec:metrics}).

\subsection{Benchmarks}\label{sec:benchmarks}
To comprehensively demonstrate our method's effectiveness and robustness, we evaluate it on four challenging benchmarks, which collectively cover a wide spectrum of real-world distortions, including documents with complete and incomplete boundaries.

\textbf{DocUNet}~\cite{ma2018docunet}.
A standard benchmark with 130 test images. Following the established protocol~\cite{ma2018docunet}, we evaluate OCR performance on its 50-image text-rich subset.

\textbf{UDIR}~\cite{feng2023deep}.
A dataset of 195 images derived from DocUNet, specifically for evaluating documents with incomplete boundaries. We adhere to its official setting, assessing OCR performance on a 70-image subset.

\textbf{WarpDoc}~\cite{xue2022fourier}.
A large-scale dataset of 1,020 images featuring six distinct distortion types (Fold, Curved, Incomplete, Random, Rotating and Perspective). For OCR evaluation, we curated a subset of 120 text-rich images by selecting 20 from each category.

\textbf{WarpDoc-Crop}.
A variant of WarpDoc created to ensure a fair comparison with methods whose architectures presume tightly cropped inputs. Each image was cropped to its document boundaries, and OCR evaluation was performed on the same 120-image subset.

\subsection{Evaluation Metrics}
\label{sec:metrics}
We evaluate our method using a suite of established metrics for image similarity and OCR accuracy. Additionally, we introduce two novel metrics designed to overcome specific limitations in current evaluation protocols.

\subsubsection{Established Evaluation Metrics}

\paragraph{Image Similarity Metrics}
Consistent with prior work~\cite{ma2018docunet, ma2022learning, wang2025axis}, we employ four metrics to assess geometric accuracy: Multi-Scale Structural Similarity (MSSIM), Local Distortion (LD), Aligned Distortion (AD) and Axis-Aligned Distortion (AAD). For the UDIR benchmark, which features incomplete boundaries, we adopt the masked versions, MSSIM-M and LD-M, as proposed by Feng et al.~\cite{feng2023deep}.

\paragraph{OCR Accuracy Metrics.}
Consistent with standard practice, we measure OCR performance using the Character Error Rate (CER)~\cite{morris2004and} and Edit Distance (ED)~\cite{levenshtein1966binary}.

\subsubsection{Proposed Enhancement Metrics}

\begin{figure}[!htbp]
    \centering
    \includegraphics[width=1.0\linewidth]{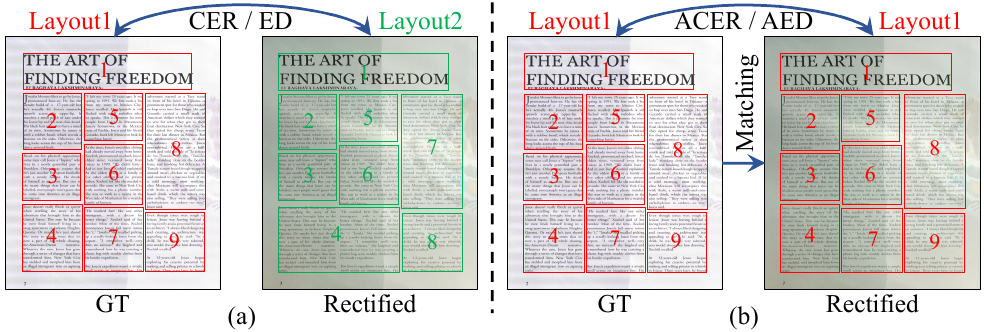}
    \caption{Comparison of OCR evaluation pipelines. (a) The conventional method vs. (b) our proposed layout-aligned method.}
    \label{fig:aocr}
\end{figure}

\paragraph{Layout-Aligned OCR Metrics (AED \& ACER)}
Conventional OCR metrics such as CER and ED can be unreliable for evaluating document rectification~\cite{das2021end, li2023layout}. Our analysis reveals that the primary source of instability is the OCR engine's layout analysis (\Cref{fig:aocr}(a)), which often misinterprets the reading order in slightly warped images. Such misinterpretation leads to scrambled text sequences and drastically inflated error metrics, imposing severe penalties that fail to reflect the true geometric rectification quality.

To decouple these factors, we introduce Layout-Aligned OCR Metrics: Aligned Character Error Rate (ACER) and Aligned Edit Distance (AED). These metrics are designed to decouple recognition errors from layout errors (\Cref{fig:aocr}(b)).  Our approach achieves this by enforcing the ground-truth (GT) reading order during evaluation, thereby isolating the assessment to the quality of geometric rectification. The procedure involves the following steps:
\begin{enumerate}
\item Extract GT Layout: Extract text blocks and their reading order from GT annotations.
\item Align Regions: Spatially align the GT text blocks with their corresponding regions in the rectified image.
\item Ordered Recognition: Perform OCR on each aligned region and concatenate the results following the GT order.
\item Calculate Metrics: Compute ACER and AED on the correctly ordered text.
\end{enumerate}

\paragraph{Masked Geometric Metrics (AAD-M \& AD-M)}
Inspired by the masked metrics (LD-M, MSSIM-M) proposed by Feng et al.~\cite{feng2023deep} for incomplete documents, we introduce AD-M and AAD-M to extend this robust evaluation principle to the AD~\cite{ma2022learning} and AAD~\cite{wang2025axis} metrics.

\begin{figure}[!htbp]
    \centering
    \includegraphics[width=1.0\linewidth]{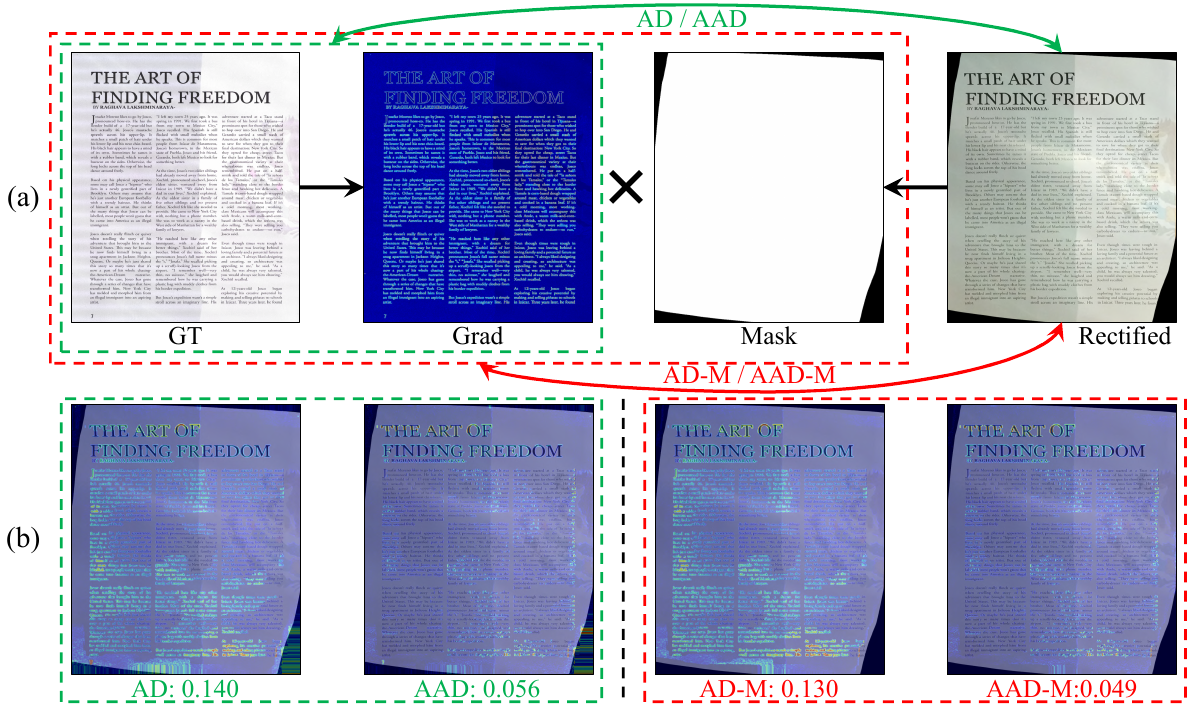}
    \caption{Comparison of the calculation process for standard AD/AAD (green box) and our proposed masked AD-M/AAD-M (red box). (a) The components used in the calculation. (b) Heat map corresponding to the calculation metrics.}
    \label{fig:AD&AAD}
\end{figure}

As illustrated in \Cref{fig:AD&AAD}(a), our method applies a foreground mask to calculation inputs—the ground truth (GT) image and their gradients (Grad)—thereby confining the evaluation to valid document regions. The resulting error heatmaps (\Cref{fig:AD&AAD}(b)) demonstrate that AD-M and AAD-M metrics successfully nullify errors in irrelevant background areas, leading to a more accurate measurement of geometric distortion.

\section{Experiments}

\subsection{Implementation Details}

\subsubsection{Training}
Following previous work~\cite{wang2025axis}, our model is trained on a composite dataset of 88k images from Doc3D~\cite{das2019dewarpnet} and 20k from UVDoc~\cite{verhoeven2023uvdoc}. All input images are resized to 712 $\times$ 488, and the cascaded L/C/F-Net architecture predicts a sparse 45 $\times$ 31 grid. We use the AdamW~\cite{loshchilov2017decoupled} optimizer with a batch size of 36 and an initial learning rate of $10^{-5}$ that follows a linear decay schedule.
Our data augmentation strategy enhances robustness by training on samples with incomplete boundaries (generated via random cropping) and superimposing documents onto diverse backgrounds from an indoor scene dataset~\cite{quattoni2009recognizing} to simulate complex, cluttered environments.
The networks are trained sequentially. The L-Net is trained for 30 epochs on data with large scale variations to specialize in localization. For geometric correction, the C-Net is trained for 30 epochs on boundary-cropped images. The F-Net is then initialized from the converged C-Net model and fine-tuned for 10 epochs to address residual errors left by the C-Net. All models were trained on 4$\times$ NVIDIA RTX 4090 GPUs.

\subsubsection{Inference}
During inference, we set the boundary extension ratio $m=0.03$ (Eq.~\ref{eq:snorm}) and the iterative stopping parameters (Alg.~\ref{alg:high_level_stop}) to $M=5$, $\theta_{thresh}=5^{\circ}$, and $\tau=0.99$.
To ensure a consistent and realistic benchmark, efficiency tests were conducted on a server with an Intel Xeon Gold 6133 CPU and an NVIDIA RTX 4090 GPU. We utilized a batch size of 1 and a standardized 4K input resolution ($4096 \times 3072$, common in mobile photography), averaging the results over 300 test images to ensure statistical stability.

\subsubsection{Evaluation}
MSSIM and its masked variant are calculated using a conventional five-level image pyramid with standard weights (0.0448, 0.2856, 0.3001, 0.2363, 0.1333); these calculations are performed in MATLAB R2024b. For all OCR-based metrics, we employ the Tesseract v5.4 engine via the PyTesseract v0.3.13~\footnote{\url{https://pypi.org/project/pytesseract/}} wrapper. To implement our proposed layout-aligned OCR metrics, we use the RAFT~\cite{teed2020raft} optical flow method for aligning the ground-truth text regions with the rectified image.
 
\subsection{Experimental Results}
To validate the efficacy of our proposed method, we conduct a comprehensive comparison against current state-of-the-art approaches through both quantitative and qualitative analyses.

\begin{table*}[!htbp]
    \centering
    \footnotesize
    \caption{Quantitative results on the DocUNet, UDIR, WarpDoc, and WarpDoc-Crop benchmarks. \textbf{Bold} and \underline{underline} denote the best and second-best results. \textbf{Para.}: Number of parameters (Millions). \textbf{FPS}: Frames Per Second (Total / Model). Higher is better (↑), lower is better (↓), which also applies to standard deviation (std).}
    \label{tab:comparison_all_benchmarks}
    
    \setlength{\tabcolsep}{1.0pt} 
    \renewcommand{\arraystretch}{0.9} 
    \begin{tabular}{@{\hspace{4pt}} l r@{\,/\,}l @{\hspace{4pt}} cccccc | cccccc @{}}
        \toprule
        
        \multirow{2}{*}{\textbf{Method}} & \multicolumn{2}{c}{\multirow{2}{*}{\textbf{Para.↓}}} & \multicolumn{6}{c}{\textbf{DocUNet benchmark}} & \multicolumn{6}{c}{\textbf{UDIR benchmark}} \\

        \cmidrule(lr){4-9} \cmidrule(lr){10-15}
        
         & \multicolumn{2}{c}{} & \textbf{MSSIM↑} & \textbf{LD↓} & \textbf{AD↓} & \textbf{AAD↓} & \textbf{CER (std)↓} & \textbf{ACER (std)↓} & \textbf{MSSIM-M↑} & \textbf{LD-M↓} & \textbf{AD-M↓} & \textbf{AAD-M↓} & \textbf{CER (std)↓} & \textbf{ACER (std)↓} \\
        \midrule
        
        DewarpNet\cite{das2019dewarpnet}   & \multicolumn{2}{c}{86.9} & 0.474 & 8.362 & 0.398 & 0.164 & 0.225 (0.210) & 0.088 (0.096) 
        & 0.361 & 18.087 & 0.754 & 0.333 & 0.313 (0.210) & 0.202 (0.154) \\
        DDCP\cite{xie2021document}$^*$     & \multicolumn{2}{c}{13.3} & 0.472 & 8.982 & 0.428 & 0.159 & 0.199 (0.195) & 0.068 (0.069)
        & 0.360 & 19.366 & 0.650 & 0.259 & 0.243 (0.204) & 0.137 (0.130) \\
        DocTr\cite{feng2021doctr}          & \multicolumn{2}{c}{26.9} & 0.509 & 7.773 & 0.369 & 0.151 & 0.185 (0.200) & 0.063 (0.061)
        & 0.381 & 18.834 & 0.735 & 0.310 & 0.289 (0.223) & 0.180 (0.144) \\
        DocGeoNet\cite{feng2022geometric}  & \multicolumn{2}{c}{24.8} & 0.504 & 7.717 & 0.381 & 0.158 & 0.187 (0.181) & 0.068 (0.083)
        & 0.381 & 18.730 & 0.742 & 0.316 & 0.287 (0.207) & 0.200 (0.150) \\
        PaperEdge\cite{ma2022learning}     & \multicolumn{2}{c}{36.6} & 0.473 & 7.967 & 0.368 & 0.121 & 0.183 (0.218) & 0.057 (0.058)
        & 0.387 & 16.792 & 0.702 & 0.218 & 0.227 (0.214) & 0.140 (0.142) \\
        FTDR\cite{li2023foreground}        & \multicolumn{2}{c}{45.2} & 0.497 & 8.416 & 0.377 & 0.151 & 0.176 (0.190) & 0.067 (0.077)
        & 0.388 & 16.415 & 0.673 & 0.286 & 0.244 (0.215) & 0.133 (0.120) \\
        DocScanner\cite{feng2025docscanner} & \multicolumn{2}{c}{\underline{8.5}} & 0.517 & 7.440 & 0.336 & 0.122 & \underline{0.150} (0.176) & 0.054 (0.056)
        & 0.388 & 18.379 & 0.679 & 0.269 & 0.265 (0.213) & 0.175 (0.141) \\
        DocTr++\cite{feng2023deep}         & \multicolumn{2}{c}{26.6} & 0.464 & 9.301 & 0.371 & 0.146 & 0.204 (0.209) & 0.061 (0.056)
        & \underline{0.452} & \underline{12.448} & \underline{0.529} & 0.216 & 0.319 (0.306) & \underline{0.115} (0.141) \\
        UVDoc\cite{verhoeven2023uvdoc}     & \multicolumn{2}{c}{\textbf{8.0}} & \textbf{0.545} & 6.827 & 0.316 & 0.125 & 0.193 (0.209) & 0.065 (0.067)
        & 0.396 & 17.260 & 0.571 & 0.239 & \underline{0.210} (0.187) & 0.134 (0.130) \\
        DocRes\cite{zhang2024docres}       & \multicolumn{2}{c}{15.2} & 0.466 & 9.371 & 0.472 & 0.199 & 0.227 (0.207) & 0.085 (0.082)
        & 0.391 & 22.757 & 0.914 & 0.368 & 0.459 (0.312) & 0.367 (0.316) \\
        AADD\cite{wang2025axis} & \multicolumn{2}{c}{2$\times$8.0} & \underline{0.543} & \underline{6.249} & \underline{0.278} & \underline{0.099} & \textbf{0.150} (0.166) & \underline{0.054} (0.059) 
        & 0.406 & 17.191 & 0.593 & \underline{0.193} & \textbf{0.189} (0.189) & 0.115 (0.127) \\
        \midrule
        Ours w/o L-Net & \multicolumn{2}{c}{2$\times$8.0} & 0.515 & 6.066 & 0.267 & 0.089 & 0.160 (0.196) & 0.049 (0.048) 
        & 0.404 & 15.348 & 0.418 & 0.142 & 0.158 (0.171) & 0.106 (0.113) \\
        Ours w/o C-Net & \multicolumn{2}{c}{2$\times$8.0} & 0.504 & 7.510 & 0.311 & 0.094 & 0.163 (0.189) & 0.050 (0.051) 
        & 0.474 & 11.274 & 0.455 & 0.139 & 0.308 (0.322) & 0.105 (0.153) \\
        Ours w/o F-Net & \multicolumn{2}{c}{2$\times$8.0} & 0.518 & 6.075 & 0.277 & 0.102 & 0.150 (0.175) & 0.054 (0.056)  
        & 0.486 & 10.228 & 0.418 & 0.146 & 0.299 (0.308) & 0.111 (0.153) \\
        \textbf{Ours}  & \multicolumn{2}{c}{3$\times$8.0} & 0.519 & \textbf{6.198} & \textbf{0.260} & \textbf{0.085} & 0.166 (0.195) & \textbf{0.049} (0.050) 
        & \textbf{0.489} & \textbf{10.345} & \textbf{0.404} & \textbf{0.126} & 0.286 (0.306) & \textbf{0.103} (0.151) \\
        
        \midrule\midrule
        
        \multirow{2}{*}{\textbf{Method}} & \multicolumn{2}{c}{\multirow{2}{*}{\textbf{FPS}↑}} & \multicolumn{6}{c}{\textbf{WarpDoc benchmark}} & \multicolumn{6}{c}{\textbf{WarpDoc-Crop benchmark}} \\
        
        \cmidrule(lr){4-9} \cmidrule(lr){10-15}
        
         & \multicolumn{2}{c}{} & \textbf{MSSIM↑} & \textbf{LD↓} & \textbf{AD↓} & \textbf{AAD↓} & \textbf{CER (std)↓} & \textbf{ACER (std)↓} & \textbf{MSSIM↑} & \textbf{LD↓} & \textbf{AD↓} & \textbf{AAD↓} & \textbf{CER (std)↓} & \textbf{ACER (std)↓} \\
        \midrule
        
        DewarpNet\cite{das2019dewarpnet}   & 0.43 & 56.4 & 0.327 & 31.075 & 0.893 & 0.364 & 0.540 (0.216) & 0.335 (0.182)
        & 0.465 & 13.303 & 0.453 & 0.206 & 0.258 (0.235) & 0.138 (0.136) \\
        DDCP\cite{xie2021document}$^*$     & \underline{0.62} & \underline{95.5} & 0.357 & 23.387 & 0.718 & 0.279 & 0.386 (0.249) & 0.201 (0.171)
        & 0.475 & 14.016 & 0.470 & 0.195 & 0.230 (0.238) & 0.116 (0.133) \\
        DocTr\cite{feng2021doctr}          & 0.51 & 28.9 & 0.394 & 27.003 & 0.777 & 0.298 & 0.380 (0.250) & 0.214 (0.172)
        & 0.521 & 11.703 & 0.347 & 0.146 & 0.145 (0.200) & 0.075 (0.113) \\
        DocGeoNet\cite{feng2022geometric}  & 0.53 & 13.6 & 0.402 & 24.701 & 0.756 & 0.307 & 0.396 (0.255) & 0.246 (0.183)
        & 0.523 & 11.602 & 0.364 & 0.157 & 0.165 (0.213) & 0.082 (0.109) \\
        PaperEdge\cite{ma2022learning}     & 0.44 & 89.6 & 0.397 & 20.333 & 0.589 & 0.195 & 0.219 (0.230) & 0.141 (0.150)
        & 0.507 & 11.271 & 0.325 & 0.110 & 0.110 (0.172) & 0.067 (0.110) \\
        FTDR\cite{li2023foreground}        & 0.48 & 7.4 & 0.368 & 26.781 & 0.755 & 0.296 & 0.438 (0.250) & 0.292 (0.212)
        & 0.519 & 11.964 & 0.367 & 0.154 & 0.173 (0.220) & 0.091 (0.124) \\
        DocScanner\cite{feng2025docscanner} & 0.53 & 21.6 & 0.392 & 27.499 & 0.638 & 0.226 & 0.267 (0.271) & 0.153 (0.163)
        & 0.527 & 11.753 & 0.316 & 0.116 & 0.123 (0.193) & 0.058 (0.100) \\
        DocTr++\cite{feng2023deep}         & -- & -- & 0.245 & 43.126 & 1.077 & 0.338 & 0.519 (0.283) & 0.360 (0.248)
        & 0.509 & 12.292 & 0.363 & 0.155 & 0.173 (0.213) & 0.097 (0.140) \\
        UVDoc\cite{verhoeven2023uvdoc}     & \textbf{0.63} & \textbf{109.0} & 0.358 & 26.277 & 0.632 & 0.256 & 0.379 (0.249) & 0.199 (0.162)
        & \underline{0.551} & 10.104 & 0.278 & 0.120 & 0.129 (0.192) & 0.066 (0.099) \\
        DocRes\cite{zhang2024docres}       & 0.27 & 20.5 & 0.501 & \underline{12.856} & 0.455 & 0.197 & 0.272 (0.227) & 0.157 (0.149)
        & 0.503 & 13.065 & 0.424 & 0.185 & 0.218 (0.214) & 0.119 (0.129) \\
        AADD\cite{wang2025axis} & 0.56 & 67.0 & \underline{0.514} & 15.642 & \underline{0.276} & \underline{0.095} & \underline{0.128} (0.206) & \underline{0.054} (0.090)
        & \textbf{0.566} & \underline{9.610} & \underline{0.218} & \underline{0.081} & \underline{0.098} (0.166) & \underline{0.041} (0.051) \\
        \midrule
        Ours w/o L-Net & 0.56 & 67.0 & 0.412 & 19.669 & 0.344 & 0.097 & 0.106 (0.178) & 0.054 (0.092)
        & 0.532 & 8.932 & 0.215 & 0.065 & 0.076 (0.143) & 0.035 (0.045) \\
        Ours w/o C-Net & 0.56 & 67.0 & 0.515 & 11.946 & 0.280 & 0.088 & 0.087 (0.152) & 0.044 (0.053)
        & 0.537 & 10.273 & 0.238 & 0.071 & 0.083 (0.151) & 0.040 (0.047) \\
        Ours w/o F-Net & 0.56 & 67.0 & 0.528 & 9.134 & 0.212 & 0.078 & 0.088 (0.162) & 0.036 (0.048)  
        & 0.538 & 8.494 & 0.205 & 0.074 & 0.097 (0.176) & 0.040 (0.053) \\
        \textbf{Ours}  & 0.51 & 48.9 & \textbf{0.534} & \textbf{9.528} & \textbf{0.201} & \textbf{0.062} & \textbf{0.079} (0.157) & \textbf{0.031} (0.038)
        & 0.544 & \textbf{8.900} & \textbf{0.198} & \textbf{0.059} & \textbf{0.073} (0.137) & \textbf{0.034} (0.044) \\
        \bottomrule
    \end{tabular}
    
    \begin{flushleft}
    \footnotesize 
    \textit{Note:} For a fair OCR metrics comparison, all methods are evaluated at the original input resolution, DDCP$^*$ outputs are upscaled accordingly. In ``Ours w/o C-Net" ablation,  the L-Net's output grid is used to perform dewarping directly, bypassing the affine transformation stage entirely. The notation $N \times 8.0$ denotes the effective parameter usage accumulation across multiple inference passes: AADD involves an additional pre-processing step ($2\times$), whereas our method employs a three-stage cascaded refinement ($3\times$).
    \end{flushleft}
\end{table*}

\subsubsection{Quantitative Comparison}
Quantitative results in \Cref{tab:comparison_all_benchmarks} demonstrate that our method establishes a new state-of-the-art across all benchmarks with substantial performance gains over previous methods. Our ACER metric has a smaller value and standard deviation compared to CER, indicating its effective stability.

Specifically, on the DocUNet benchmark, our method improves the AAD metric from 0.099 to 0.085, with a corresponding reduction in the OCR-related metric, ACER, from 0.054 to 0.049. This performance disparity is even more pronounced on the UDIR dataset, where our method consistently outperforms all competitors. The overall alignment metric, AD-M, is significantly reduced from 0.529 to 0.404, while the axis alignment metric, AAD-M, drops from 0.193 to 0.126. This significantly enhances the rectification accuracy for document images with incomplete boundaries. Notably, this benchmark highlights the stability of our proposed ACER metric: while conventional CER paradoxically assigns poor scores to geometrically superior methods like ours and DocTr++~\cite{feng2023deep}, ACER aligns correctly with the visual and geometric results, providing a more faithful evaluation.

On the WarpDoc benchmark, our method reduces the AAD metric from 0.095 to 0.062, demonstrating enhanced rectification accuracy and superior robustness amidst complex background interference, which provides strong support for real-world applications. Notably, on the cropped version of the benchmark (WarpDoc-Crop), most prior methods show improved performance, as the removal of significant background noise simplifies the rectification task. In contrast, our method's performance remains remarkably consistent between the uncropped and cropped versions. This consistency underscores the inherent robustness of our approach to variations in background interference and object scale. On this subset, our method again surpasses the previous best, lowering the AAD metric from 0.081 to 0.059.

Ablation studies on each benchmark confirm the synergistic nature of our cascaded networks. The F-Net delivers the most significant gains on datasets like DocUNet and WarpDoc-Crop by refining fine-grained details. For more challenging datasets with diverse distortions, such as UDIR and WarpDoc, the L-Net's robust initial localization is indispensable. By first mitigating the primary perspective distortion, it allows the C-Net to more effectively resolve the remaining complex shape deformations. This clear division of labor is key to our unified strategy, enabling it to achieve state-of-the-art results across all benchmarks.

\subsubsection{Qualitative Comparison}
\begin{figure*}[!htbp]
    \centering
    \includegraphics[width=1.0\linewidth]{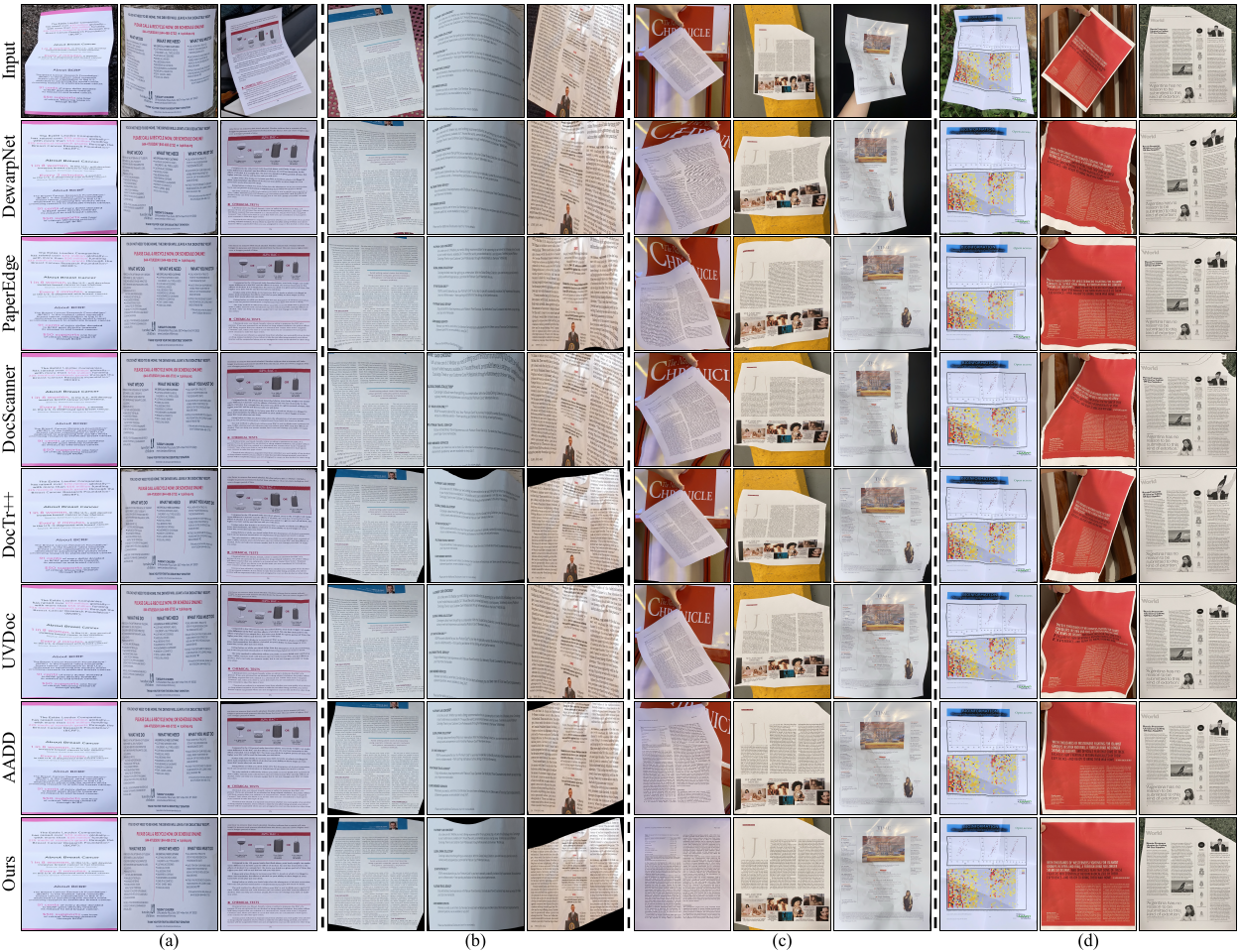}
    \caption{Qualitative comparison with prior works on four benchmarks: (a) DocUNet, (b) UDIR, (c) WarpDoc, and (d) WarpDoc-Crop.}
    \label{fig:compare_qualitative_all}
\end{figure*}

A qualitative comparison across the four benchmarks is presented in~\Cref{fig:compare_qualitative_all}. As shown in \Cref{fig:compare_qualitative_all}(a) for the DocUNet benchmark, our method excels at mitigating internal content distortions, corroborating its superior AAD metric reported in \Cref{tab:comparison_all_benchmarks}. The iterative fine-tuning process further minimizes content warping and enhances text readability. On the UDIR benchmark, our method accurately identifies and rectifies the valid document region, ensuring precise axis alignment. In contrast to DocTr++~\cite{feng2023deep}, the reduction in content distortion is particularly striking, demonstrating a significant improvement on this benchmark.
As visualized on WarpDoc in \Cref{fig:compare_qualitative_all}(c), previous methods generally struggle with these distortions, especially in the presence of complex backgrounds, whereas our approach remains robust. Consistent with the quantitative analysis in \Cref{tab:comparison_all_benchmarks}, while prior methods show slight improvements on the cropped version in~\Cref{fig:compare_qualitative_all}(d), our method, leveraging its accurate localization, achieves even lower content distortion. This underscores the high robustness of our method across diverse testing conditions.

\subsubsection{Comparison with Commercial Applications} 
To assess the practical effectiveness of our approach, we compared it with leading commercial applications. For each commercial system, we obtained rectification results via its official API, ensuring the evaluation reflects production-level performance.

\begin{savenotes} 
\begin{table}[!htbp]
\centering
\caption{Quantitative comparison with leading commercial applications on the UDIR benchmark. \textbf{Bold} indicates the best performance. ($\uparrow$) Higher is better; ($\downarrow$) Lower is better.}
\label{tab:comparison_commercial}
\setlength{\tabcolsep}{1.0pt} 
\renewcommand{\arraystretch}{1.0} 
\begin{tabular}{@{} l cccccc @{}}
\toprule
\textbf{Method} & \textbf{MSSIM-M$\uparrow$} & \textbf{LD-M$\downarrow$} & \textbf{AD-M$\downarrow$} & \textbf{AAD-M$\downarrow$} & \textbf{CER (std)$\downarrow$} & \textbf{ACER (std)$\downarrow$} \\
\midrule
Youdao\footnote{\url{https://ai.youdao.com/}\label{fn:youdao}} & 0.334 & 23.931 & 0.993 & 0.452 & {0.307}{(0.242)} & {0.190}{(0.175)} \\
Baidu\footnote{\url{https://cloud.baidu.com/}\label{fn:baidu}}   & 0.378 & 17.897 & 0.578 & 0.201 & {0.227}{(0.223)} & {0.111}{(0.132)} \\
TextIn\footnote{\url{https://www.textin.com/}\label{fn:textin}} & 0.456 & 12.889 & 0.455 & 0.152 & \textbf{{0.201}{(0.210)}} & {0.108}{\textbf{(0.128)}} \\
\textbf{Ours} & \textbf{0.489} & \textbf{10.345} & \textbf{0.404} & \textbf{0.126} & {{0.286}{(0.306)}} & \textbf{0.103}(0.151) \\
\bottomrule
\end{tabular}

\begin{flushleft}
\footnotesize
\textit{Note: All commercial baselines were accessed on July 29, 2025.}
\end{flushleft}

\end{table}
\end{savenotes}

The quantitative comparison on the challenging UDIR~\cite{feng2023deep} benchmark is presented in~\Cref{tab:comparison_commercial}. Our method consistently outperforms all commercial competitors across the key geometric metrics. Specifically, compared to \textit{TextIn}, which exhibits the best performance among the commercial tools, our method achieves a significant improvement in geometric rectification accuracy, reducing the AAD-M metric from 0.152 to 0.126. 

Notably, this comparison is conducted under conditions inherently disadvantageous to our method. Commercial solutions benefit from massive proprietary datasets and extensive engineering optimizations accumulated over years of deployment. In contrast, our method relies solely on public datasets and a transparent optimization strategy. Despite this data disadvantage, our model still surpasses the commercial state-of-the-art, demonstrating superior efficiency and generalization capability.

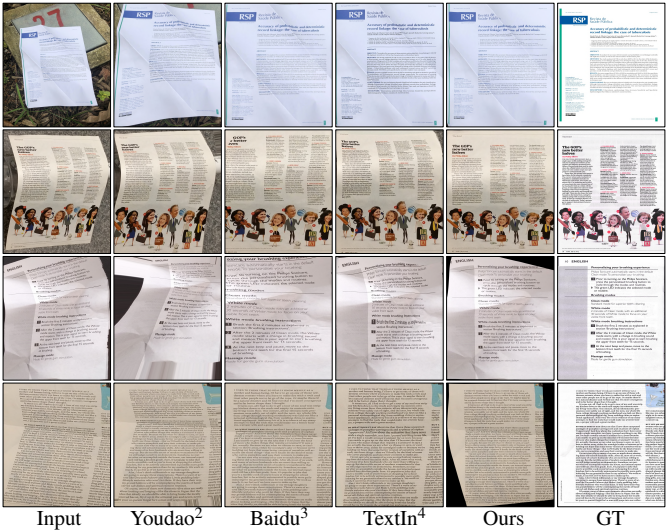
\begin{figure}[!htbp]
    \centering
    \begin{overpic}[width=1.0\linewidth]{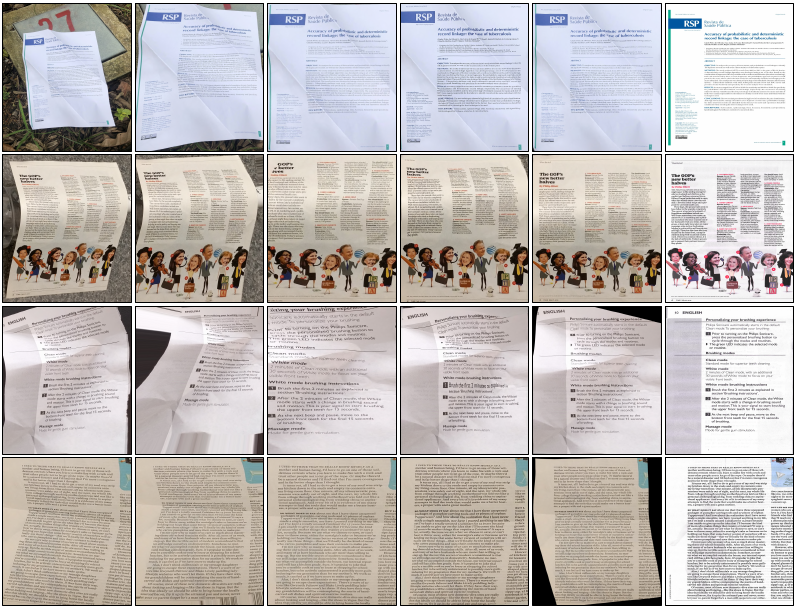}
        \put(4, -2.5){ \footnotesize Input}
        \put(18, -2.5){ \footnotesize Youdao\textsuperscript{\ref{fn:youdao}}}
        \put(36, -2.5){ \footnotesize Baidu\textsuperscript{\ref{fn:baidu}}}
        \put(53, -2.5){ \footnotesize TextIn\textsuperscript{\ref{fn:textin}}}
        \put(71, -2.5){ \footnotesize Ours}
        \put(88, -2.5){ \footnotesize GT}
    \end{overpic}
    \caption{Qualitative comparison of our method against leading commercial products on challenging images with various boundary conditions.}
    \label{fig:compare_product}
\end{figure}

In addition to quantitative metrics, we visually compare our method with these commercial products in \Cref{fig:compare_product}, covering three distinct challenging scenarios. In the first row, characterized by complex background textures, commercial products fail to effectively localize the document boundaries due to visual clutter. When facing severe geometric distortions (Row 2), these tools exhibit limitations in effectively rectifying the warped content areas. Moreover, for documents with incomplete boundaries (Rows 3-4), commercial baselines struggle to infer the correct structure, producing erroneous geometric shapes due to missing visual cues. In sharp contrast, our method demonstrates remarkable robustness across all these conditions, accurately recovering the document geometry where other solutions falter.

\subsubsection{Efficiency Comparison}
As detailed in Table \ref{tab:comparison_all_benchmarks}, we conduct a comprehensive evaluation of computational efficiency in terms of model parameters (Para.) and inference speed (FPS).

\textbf{Model Complexity.} 
Adhering to lightweight design principles~\cite{verhoeven2023uvdoc,wang2025axis}, our framework utilizes a compact backbone. Although the cascaded strategy involves three refinement stages, the total parameter count remains highly efficient. Specifically, with a total of $24.0\text{M}$ parameters ($3 \times 8.0\text{M}$), our model is significantly smaller than prior heavy-weight single-stage approaches, ensuring high storage efficiency without compromising geometric accuracy.

\textbf{Execution Speed.} We evaluate speed using two metrics: \textit{Model FPS} (pure GPU inference) and \textit{Total FPS} (end-to-end including I/O). Despite involving three sequential passes, our method achieves a high Model FPS of $48.9$, surpassing most geometric rectification methods. In real-world scenarios, our Total FPS of $0.51$ remains competitive, striking an optimal balance between high-precision rectification and practical execution speed.

\textbf{Bottleneck Analysis.} A critical insight is that the bottleneck lies in the CPU-bound data pipeline rather than neural inference. Specifically, the speed gap between the fastest and slowest models on the GPU is substantial ($\sim14.7\times$). Yet, this disparity shrinks dramatically to only $\sim2.3\times$ in Total FPS across the benchmark. For our method, GPU inference ($\sim20\text{ms}$) accounts for only $\sim4\%$ of the total latency. A similar low inference ratio is observed across other methods, confirming that high-resolution image I/O and re-sampling constitute the primary computational costs.

\textbf{Future Optimization.} Consequently, further model compression would yield negligible gains in system-level latency. Future research for practical deployment should prioritize optimizing the data processing pipeline (e.g., parallelized image I/O) rather than solely focusing on network architecture.

\subsection{Ablation Studies}
We conducted several ablation studies on the WarpDoc benchmark to validate the design choices and contributions of each key component in our framework. The results, summarized in \Cref{tab:ablation_warpdoc_compact}, are analyzed in the following sections.

\begin{table}[!htbp]
\centering
\caption{Ablation results on the \textbf{WarpDoc} benchmark. Bold denotes the best result in each subgroup.}
\label{tab:ablation_warpdoc_compact}
\setlength{\tabcolsep}{1pt}
\sisetup{detect-weight, mode=text}
\renewrobustcmd{\bfseries}{\fontseries{b}\selectfont}
\renewrobustcmd{\boldmath}{}
\renewcommand{\arraystretch}{0.85} 
\newcommand{\meanstd}[2]{#1\,{\footnotesize(#2)}}

\begin{tabular}{@{}
  l @{\hskip 3pt}|@{\hskip 3pt}            
  S[table-format=1.3] @{\hskip 6pt}        
  S[table-format=2.3] @{\hskip 6pt}        
  S[table-format=1.3] @{\hskip 6pt}        
  S[table-format=1.3] @{\hskip 10pt}       
  l                   @{\hskip 10pt}       
  l                                       
  @{}}
\toprule
\textbf{Setting} & {\textbf{MSSIM↑}} & {\textbf{LD↓}} & {\textbf{AD↓}} & {\textbf{AAD↓}} & {\textbf{CER (std)↓}} & {\textbf{ACER (std)↓}} \\
\midrule
\multicolumn{7}{@{}l}{\emph{(a) Geometric transform (GT)}}\\
\texttt{GT:S}         & 0.530 & 9.798 & 0.214 & 0.062 & \meanstd{0.072}{0.143} & \meanstd{0.031}{0.038} \\
\texttt{GT:P}         & 0.531 & 9.839 & 0.219 & 0.062 & \meanstd{0.083}{0.158} & \meanstd{0.030}{0.040} \\
\texttt{GT:\textbf{A}} & \bfseries 0.534 & \bfseries 9.528 & \bfseries 0.201 & \bfseries 0.062 & \bfseries \meanstd{0.079}{0.157} & \bfseries \meanstd{0.031}{0.038} \\

\addlinespace[2pt]
\multicolumn{7}{@{}l}{\emph{(b) Iteration strategies (IS)}}\\
\texttt{IS:0}         & 0.528 & \bfseries 9.135 & 0.212 & 0.078 & \meanstd{0.088}{0.162} & \meanstd{0.036}{0.048} \\
\texttt{IS:1}         & \bfseries 0.536 & 9.474 & 0.205 & 0.062 & \meanstd{0.081}{0.161} &  \meanstd{0.033}{0.041} \\
\texttt{IS:2}         & 0.533 & 9.962 & 0.215 & 0.062 & \meanstd{0.071}{0.135} & \meanstd{0.034}{0.037} \\
\texttt{IS:3}         & 0.529 & 10.404 & 0.229 & 0.063 & \meanstd{0.068}{0.136} & \meanstd{0.033}{0.038} \\
\texttt{IS:4}         & 0.526 & 10.714 & 0.245 & 0.063 & \meanstd{0.074}{0.154} & \meanstd{0.033}{0.040} \\
\texttt{IS:5}         & 0.524 & 11.071 & 0.259 & 0.065 & \bfseries \meanstd{0.067}{0.131} & \meanstd{0.035}{0.040} \\
\texttt{IS:\textbf{A}} & 0.534 & 9.528 & \bfseries 0.201 & \bfseries 0.062 & \meanstd{0.079}{0.157} & \bfseries \meanstd{0.031}{0.038} \\

\addlinespace[2pt]
\multicolumn{7}{@{}l}{\emph{(c) Affine loss (AL)}}\\
\texttt{AL:w/o}       & 0.536 & 10.150 & 0.211 & 0.064 & \bfseries \meanstd{0.070}{0.141} & \meanstd{0.032}{0.040} \\
\texttt{AL:\textbf{w}} & \bfseries 0.534 & \bfseries 9.528 & \bfseries 0.201 & \bfseries 0.062 & \meanstd{0.079}{0.157} & \bfseries \meanstd{0.031}{0.038} \\
\bottomrule
\end{tabular}
\vspace{4pt}
\begin{flushleft} 
\footnotesize
\textit{Note:} Settings are abbreviated as follows. (a) Geometric transforms: S = Similarity, P = Perspective, A = Affine (our proposal). (b) Iteration strategies: 0-5 = fixed number of iterations, A = Adaptive (our proposal). (c) Affine loss: w/o = without, w = with.
\end{flushleft}
\end{table}

\subsubsection{Effectiveness of the Affine Transformation}
Our analysis of different geometric transformations for the localization network, presented in \Cref{tab:ablation_warpdoc_compact}(a), confirms that our proposed affine transformation achieves the best performance. While a perspective transformation offers stronger initial unwarping, it often introduces artifacts because wrinkled documents are not perfectly planar. Conversely, the more constrained similarity transformation lacks the necessary degrees of freedom. The affine transform strikes an optimal balance: it is powerful enough to correct common deformations while preserving crucial properties like parallelism, making it ideally suited for document rectification.

\subsubsection{Effectiveness of Adaptive Iteration}
The ablation on iteration count, shown in \Cref{tab:ablation_warpdoc_compact}(b), reveals a clear trade-off. Initial iterations progressively reduce the AAD score, but further iterations lead to diminishing returns and noise accumulation, eventually degrading performance. Our adaptive iteration strategy solves this by dynamically terminating the process at the optimal point, thereby preventing over-correction and securing the best possible AAD score.

\subsubsection{Effectiveness of the Affine Loss}
The inclusion of our proposed Affine Loss yields a noticeable performance gain, as shown in the ablation results in~\Cref{tab:ablation_warpdoc_compact}(c). This improvement stems from the loss providing more effective supervision for the model's global localization learning, which directly enhances the overall rectification accuracy.

\subsubsection{Robustness of the Layout-Aligned OCR Metrics}
\begin{figure}[!htbp]
    \centering
    \includegraphics[width=1.0\linewidth]{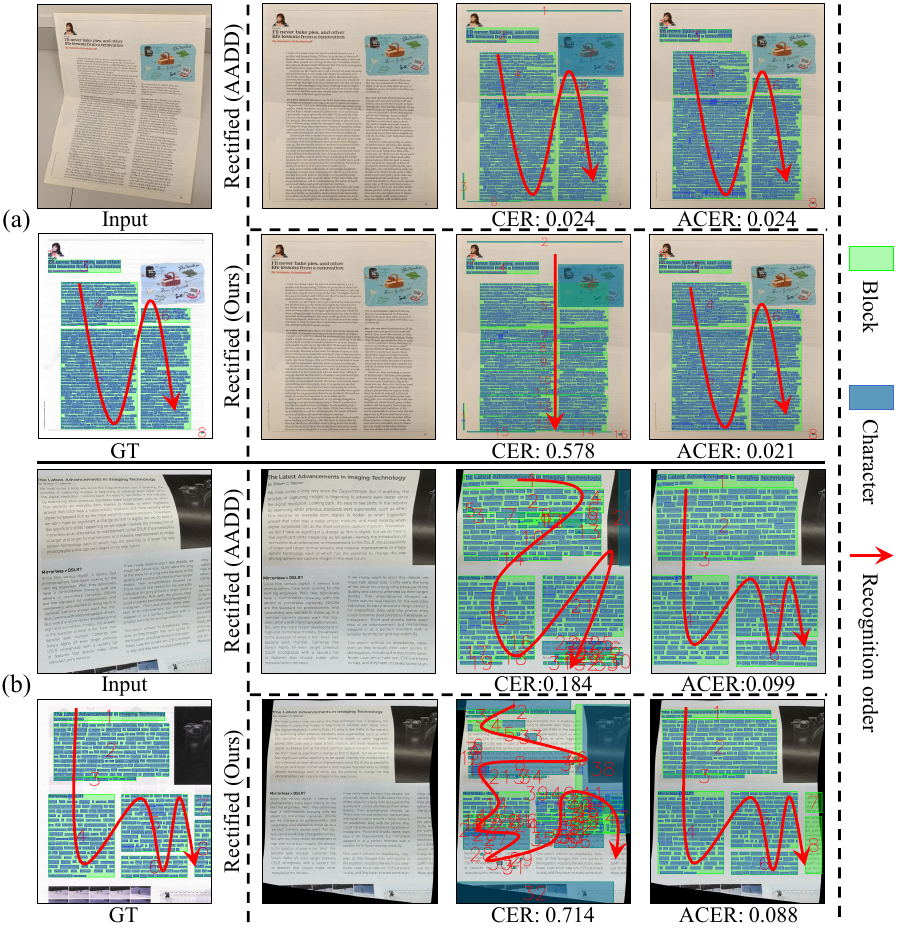}
    \caption{Comparison of Failure Cases in Conventional OCR Evaluation. (a) DocUNet~\cite{ma2018docunet} example: Similar rectified images yield large, inconsistent errors due to reading order issues. (b) UDIR~\cite{feng2023deep} example: Dark borders lead to a failure in text detection, with large areas being ignored.}
    \label{fig:LAYOUTOCR}
\end{figure}

In \Cref{fig:LAYOUTOCR}, we show failure cases of the conventional CER metric in evaluating rectified document images. As can be seen, CER fails to reflect the true dewarping quality due to the OCR engine's independent layout parsing errors, such as inferring the wrong reading order (a) or failing to detect entire text blocks (b). In contrast, our proposed ACER metric is more robust and reliable because it isolates the evaluation of rectification from these layout analysis failures.

\section{Limitations and Future Work}\label{sec:limitations}

\begin{figure}[!htbp]
    \centering
    \includegraphics[width=1.0\linewidth]{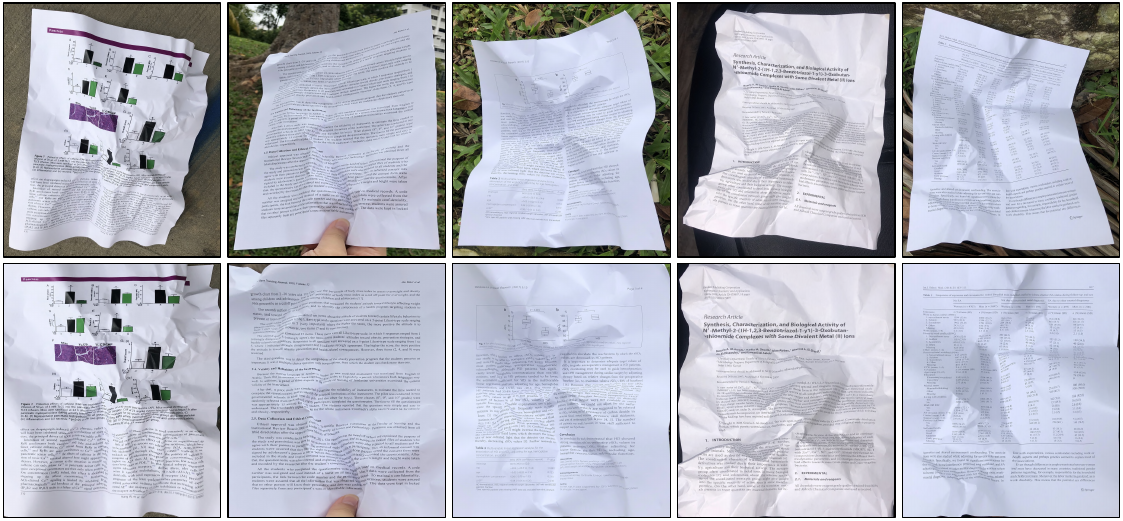}
    \caption{Failure cases on documents with severe, complex wrinkles. Top row: input images. Bottom row: our method's rectified outputs.}
    \label{fig:limit}
\end{figure}

While our method achieves state-of-the-art performance on existing benchmarks, it has limitations when faced with documents exhibiting extremely severe and complex folds, such as those in the ``random'' category of the WarpDoc~\cite{xue2022fourier} benchmark. As shown in \Cref{fig:limit}, although our approach significantly improves the document's readability, it falls short of achieving a perfectly flat rectification.

We attribute this limitation to two primary factors. First, the scarcity of training data containing such extreme, high-frequency distortions prevents the network from learning these patterns. Second, our method, like other current data-driven approaches, is fundamentally limited by the inherent ambiguity of single-view reconstruction.

For future work, we believe two avenues are promising. The first is the development of more diverse and challenging datasets that include these complex distortion types. The second, more transformative direction, is to move beyond single-view rectification. Recent advances in multi-view 3D modeling, such as 3D Gaussian Splatting~\cite{kerbl20233d}, have shown remarkable success in object reconstruction. We posit that leveraging multi-view information could fundamentally overcome the inherent ambiguities of single-image correction and enabling near-perfect document rectification in the future.

\section{Conclusion}
This paper addresses the rectification of arbitrary document images, which suffer from a complex mix of distortions. We propose an adaptive, cascaded framework that progressively corrects these distortions in a coarse-to-fine process. An initial affine transformation normalizes the input, enabling subsequent coarse and fine-tuning networks to effectively resolve geometric and content warps.
To address unreliable standard evaluations, we also introduce two key metrics: Layout-Aligned OCR metrics (ACER/AED) decouple rectification quality from OCR layout errors, while Masked geometric metrics (AD-M/AAD-M) enable accurate assessment of documents with incomplete boundaries, thereby collectively establishing a more reliable evaluation protocol for the field.
Extensive experiments show our method achieves state-of-the-art results on multiple challenging benchmarks. It demonstrates high robustness in real-world scenarios, outperforming prior methods and commercial systems. Our work validates the effectiveness of this staged approach and provides a solid foundation for future research in document digitization.

{
\footnotesize
\bibliographystyle{IEEEtran}
\bibliography{main}
}

\begin{IEEEbiography}[{\includegraphics[width=1in,height=1.25in,clip,keepaspectratio]{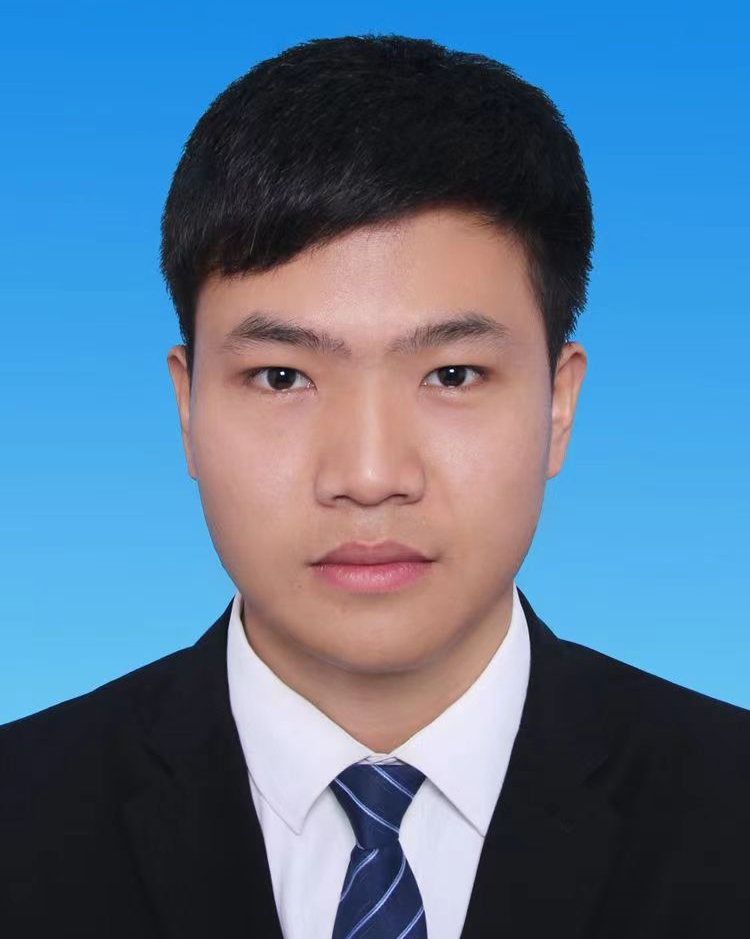}}]{Chaoyun Wang}
received the M.S. degree in control science and engineering from Harbin Engineering University, Harbin, China, in 2021. He is currently working toward the PhD degree at the Institute of Artificial Intelligence and Robotics, Xi’an Jiaotong University, Xi’an, China. His research interests include intelligent graphics and computer vision.
\end{IEEEbiography}

\vspace{-3em}

\begin{IEEEbiography}[{\includegraphics[width=1in,height=1.25in,clip,keepaspectratio]{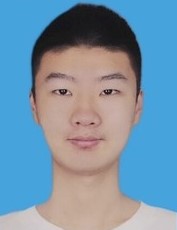}}]{Quanxin Huang}
received the B.S. degree in Artificial Intelligence from Xi'an Jiaotong University, Xi'an, China, in 2024. He is currently pursuing the M.S. degree at the same university. His research interests include geometry processing and computer vision.
\end{IEEEbiography}
\vspace{-3em}

\begin{IEEEbiography}[{\includegraphics[width=1in,height=1.25in,clip,keepaspectratio]{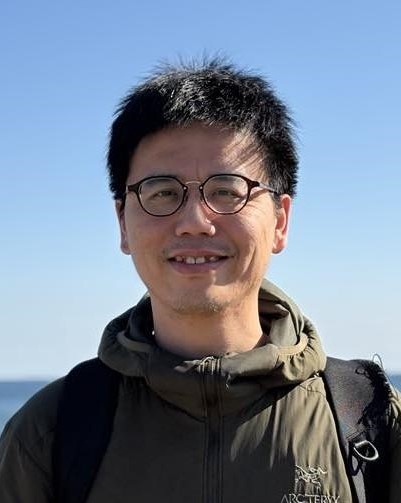}}]{I-Chao Shen} is an assistant professor at the Department of Computer Science, The University of Tokyo. He received the BBA and MBA degrees in information management from National Taiwan University in 2009 and 2011, respectively, and the PhD degree with Computer Graphics Group from National Taiwan University in 2020. Prior to his current position, he was a postdoctoral fellow and project assistant professor at the University of Tokyo. His research interests include computer graphics, machine learning, and visual computing.
\end{IEEEbiography}
\vspace{-3em}

\begin{IEEEbiography}[{\includegraphics[width=1in,height=1.25in,clip,keepaspectratio]{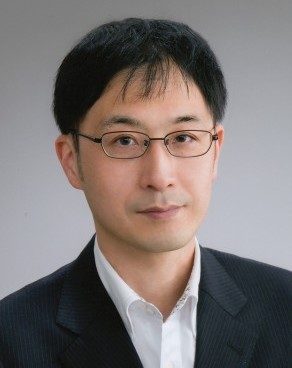}}]{Takeo Igarashi} is a professor at the Department of Creative Informatics, The University of Tokyo. His research interests are in user interfaces and interactive computer graphics. He has received the ACM SIGGRAPH 2006 Significant New Researcher Award, CHI 2019 Academy Award, and ACM UIST 2019 Lasting Impact Award. He served as a program co-chair for UIST 2013, general co-chair for UIST 2016, technical papers chair for SIGGRAPH ASIA 2018, technical program co-chair for ACM CHI 2021, and conference chair for SIGGRAPH ASIA 2024.
\end{IEEEbiography}
\vspace{-3em}

\begin{IEEEbiography}[{\includegraphics[width=1in,height=1.25in,clip,keepaspectratio]{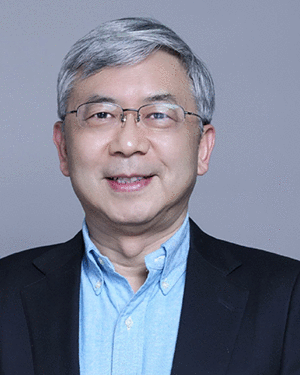}}]{Nanning Zheng} (Life Fellow, IEEE) received the graduate degree from the Department of Electrical Engineering, Xi’an Jiaotong University (XJTU), in 1975, the ME degree in information and control engineering from Xi’an Jiaotong University, in 1981, and the PhD degree in electrical engineering from Keio University, in 1985. He is currently a professor and the director of the Institute of Artificial Intelligence and Robotics of Xi’an Jiaotong University. His research interests include computer vision, pattern recognition, and hardware implementation of intelligent systems. Since 2000, he has been the Chinese representative on the Governing Board of the International Association for Pattern Recognition. He became a member of the Chinese Academy of Engineering in 1999.
\end{IEEEbiography}
\vspace{-3em}

\begin{IEEEbiography}[{\includegraphics[width=1in,height=1.25in,clip,keepaspectratio]{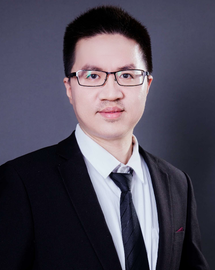}}]{Caigui Jiang} is a Professor at the Institute of Artificial Intelligence and Robotics, Xi'an Jiaotong University (XJTU), China. He received the B.S. and M.S. degrees from XJTU in 2008 and 2011, respectively, and the Ph.D. degree from King Abdullah University of Science and Technology (KAUST) in 2016. Prior to joining XJTU, he was a Research Scientist and Postdoc at the Visual Computing Center (VCC), KAUST; ICSI, UC Berkeley; and the Max Planck Institute for Informatics. His research interests include computer graphics, geometric modeling, intelligent vehicles, and robotics.
\end{IEEEbiography}

\end{document}